\documentclass[11pt,a4paper]{article}
\usepackage[noblocks]{authblk}
\usepackage{acl2015}
\usepackage{graphicx,subfigure,url,listings,verbatim} 

\begin{document}
\setcounter{Maxaffil}{2}
\title{Topic Stability over Noisy Sources}

\author[1]{Jing Su}
\author[1]{Ois\'{i}n Boydell}
\author[2]{Derek Greene}
\author[1]{Gerard Lynch}
\affil[1]{Centre for Applied Data Analytics Research}
\affil[2]{School of Computer Science \& Informatics}
\affil[ ]{University College Dublin}
\affil[ ]{\texttt{\{jing.su, oisin.boydell, derek.greene, gerard.lynch\}@ucd.ie}}

\renewcommand\Authands{ and }

\maketitle
\begin{abstract}

Topic modelling techniques such as LDA 
have recently been applied to speech transcripts and OCR output. These corpora may contain noisy or erroneous texts which may undermine topic stability. Therefore, it is important to know how well a topic modelling algorithm will perform when applied to noisy data. In this paper we show that different types of textual noise will have diverse effects on the stability of different topic models. From these observations, we propose guidelines for text corpus generation, with a focus on automatic speech transcription. We also suggest topic model selection methods for noisy corpora.

\end{abstract}


\section{Introduction}
\label{sec:intro}

Topic modelling techniques are widely applied in text retrieval tasks. Such techniques have been previously applied to news sources \cite{bib:Newman2006AET} , OCR \cite{bib:Tamura2013}, blogs \cite{bib:Yokomoto2012} etc. in which the quality of the source text is high with low error rates (missing, misspelled, or incorrect terms or phrases). However with the improvements in terms of accuracy and the reduction in the cost of automatic speech transcription and optical character recognition (OCR) technologies, the range of sources that topic modelling can now be applied to is growing. One artefact of such new text sources is their inherent noise.
In speech to text transcriptions, humans in general manage a WER of 2\% to 4\% \cite{bib:Fiscus2007}. When transcribing with a vocabulary size of  200, 5000 and 100000, the word error rates are 3\%, 7\% and 45\% respectively. The best accuracy for broadcast news transcription 13\% \cite{bib:Pallet2003}, but this drops below 25.7\% in conference transcription and gets  worse in casual conversation \cite{bib:Fiscus2007}. These records show that the difficulty of automatic speech recognition rises with vocabulary size, speaker dependency and level of crosstalk.

Noise aside, many of these newly available sources contain rich and valuable information that can be analysed through topic modelling. For example, automatic speech transcription applied to call centre audio recordings is able to capture a level of detail that is otherwise unavailable unless the call audio is manually reviewed which is infeasible for large call volumes. In this case topic modelling can be applied to transcribed text to extract the key issues and emerging topics of discussion.

In this study we propose a method for simulating various types of transcription errors. We then test the robustness of a popular topic modelling algorithm, Latent Dirichlet Allocation (LDA) using a topic stability measure introduced by \newcite{bib:Greene14topics} over a variety of corpora.

\section{Topic Modelling and Metrics} 
\label{sec:methods}

Blei et al. \cite{bib:Blei2003LDA} introduced Latent Dirichlet Allocation (LDA) as a generative probabilistic model for text corpora. LDA regulates the probabilistic distributions between document, topic and word and it is an unsupervised learning model. 

For the evaluation of topic models, we follow the approach by \newcite{bib:Greene14topics} for measuring topic model agreement . 

We can denote a topic list as $S=\{R_1,...,R_k\}$, where $R_i$ is a topic with rank $i$. An individual topic can be described as $R=\{T_1,...,T_m\}$, where $T_l$ is a term with rank $l$ belong to the topic. Jaccard index  \cite{bib:Jaccard1912} compares the number of identical items in two sets, but it neglects ranking order. Average Jaccard (AJ) similarity is a top-weighted version of the Jaccard index used to accommodate ranking information. AJ calculates the average of the Jaccard scores between every pair of subsets in two lists.
Based on AJ, we can evaluate the agreement of two sets of ranked lists (topic models). The topic model agreement score between $S_1$ and $S_2$ is a mean value of the top similarity scores between each cross pair of $R$. The agreement score is solved using the Hungarian method \cite{bib:Kuhn1955} and is constrained in the range [0,1], where a perfect match between two identical $k$-way ranked sets results in 1 and a score 0 for non-overlapping sets.
\cite{bib:Greene14topics}

\section{Datasets}
\label{sec:data}
In this paper, we explore two datasets \textit{bbc} and \textit{wikilow} \cite{bib:Greene14topics} with different document size and corpus size. The \textit{bbc} corpus includes general BBC news articles. This corpus contains 2225 documents in 5 topics and it uses 3121 terms.  The \textit{wikilow} corpus is a subset of Wikipedia and articles are labeled with fine-grained WikiProject sub-groups. There are totally 4986 documents in 10 topics and it uses 15411 terms. In both datasets the topics consist of distinct vocabularies which we expect LDA to detect. For example, the topics in \textit{bbc} datasets are \textit{business}, \textit{entertainment}, \textit{politics}, \textit{sport} and \textit{technology}.

\subsection{Textual Noise}
We artificially introduce noise into text to investigate the performance of topc modelling over naturally noisy sources.
We measure noise using word error rate (WER), a common metric for measuring speech recognition accuracy. Moreover, WER has been used as a salient metric in speech quality analytics \cite{bib:Saon2006} and spoken dialogue system \cite{bib:Cavazza2001}. In Equation \ref{eq:wer} WER is defined as the fraction between the sum of the number of substitutions S, the number of deletions D, the number of insertions and the number of terms in reference N.
  
\begin{equation}
\label{eq:wer}
WER=\frac{S+D+I}{N}
\end{equation}

\begin{table}[t]
\centering
\caption{An example of $Metaphone$ $ replacement$  in $bbc$ corpus}
\label{table:metaphone}
\begin{small}
\begin{tabular}{|l|l|}
\hline
original corpus & replaced corpus \\ 
\hline

We are hoping to  & We are hoping to  \\ 
understand the & understand the   \\
\textbf{creative}  \textbf{industry}...   & \textbf{Cardiff} \textbf{induced} ...  \\
\hline
\end{tabular}
\end{small}
\end{table}

The experiments investigate the robustness of topic models against each type of noise, and at which noise levels the output of a topic model is consistent with that of the original corpus. \textit{Deletion} noise is introduced by randomly removing a portion of text in the corpus. The proportion of deletion ranges from 0\% to 50\% and the term selection is based on uniform distribution. \textit{Insertion} is introduced by adding a portion (0\% to 50\%) of frequent terms from a list of frequent English words with 7726 entries\footnote{http://ucrel.lancs.ac.uk/bncfreq/flists.html}. The probability of sampling of a certain term from the list is based on the term frequency. 

\subsection{Metaphone Replacement} 
\label{sec:metaphone}

\begin{table}[t]
\centering
\begin{small}
\caption{Double Metaphone dictionary where terms are ranked with descending frequencies}
\label{table:dbl_meta}
\begin{tabular}{|l|l|}
\hline
Metaphone & matching terms \\ 
\hline
\textbf{ANTS} &  \textbf{industry}, units, induced, wound, ...\\
\hline
\textbf{KRTF} &   grateful, \textbf{creative}, Cardiff, ...\\
\hline
\end{tabular}
\end{small}
\end{table}

We simulate speech recognition errors using Metaphone, a phonetic algorithm for indexing English words by their pronunciation \cite{bib:Philips1990}. 
Here we use the Double Metaphone \cite{bib:Black2014} algorithm in replacement and the replacement is on a one-to-one basis. This may not simulate the full range of errors produced by ASR systems, in which the substitution may be a one-to-many or many-to-one\footnote{ e.g. \textit{recognise speech} to \textit{wreck a nice beach}.}  mapping, but it was deemed sufficient for the current experiments.

In this study we map Metaphone codes to frequent English words (examples in Table \ref{table:dbl_meta}). Then in a given text document, we randomly select X percent terms and replace each by a term in the Metaphone map. The candidate terms sharing the same metaphone symbol are selected based on term frequencies. A frequent term has higher probability to be selected over a rare term (see Table \ref{table:metaphone}).

\section{Experiments}
\label{sec:exp}

In our experiments with LDA, we aim to test the topic stability over different levels of noise and different numbers of topics. In order to produce consistent and repeatable results where each noise generation method relies on a degree of randomness with word deletion, insertion or substitution we generate multiple copies of each modified corpus using 5 random seeds. Similarly we perform 5 runs of each Mallet LDA \cite{bib:McCallum2002} topic model as the algorithm initial state is determined by a random seed. LDA hyperparameters are defined with default values, and each topic is represented by the top 25 terms. The final stability score on each level is a mean value of a number of runs with fixed seeds. 

\subsection{LDA output}
\label{sec:lda_output}

\begin{figure*}[htp]
\label{fig:bbc}
\centering
 \includegraphics[width=0.95\textwidth]{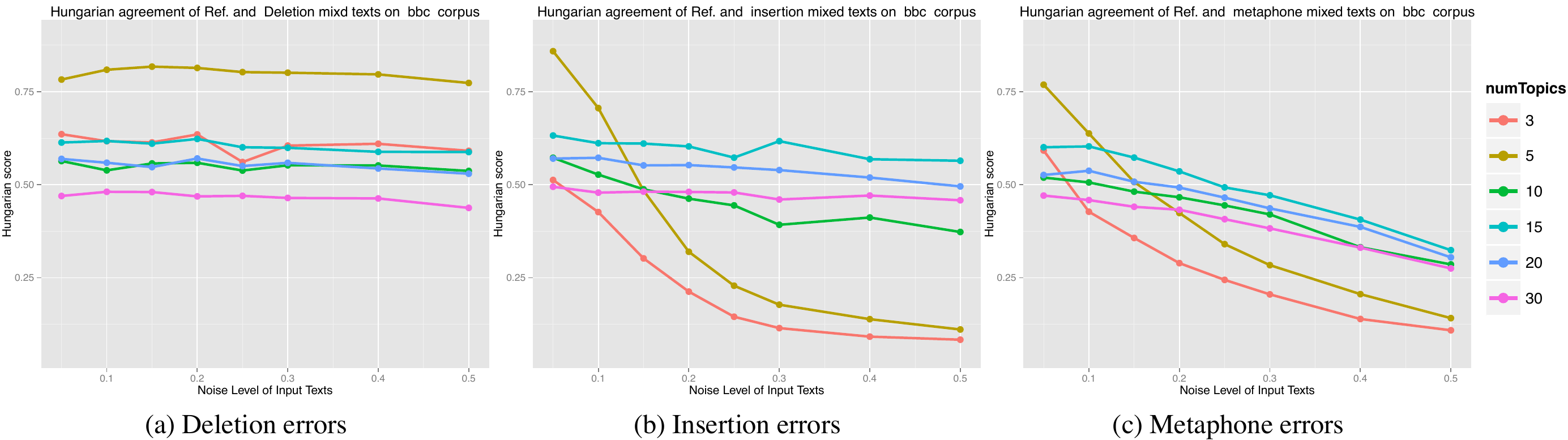}
 \caption{LDA Hungarian scores against noise levels in \textit{bbc} corpus (5 topics in reference)}
\end{figure*}

\begin{figure*}[htp]
\label{fig:wikilow}
\centering
 \includegraphics[width=0.95\textwidth]{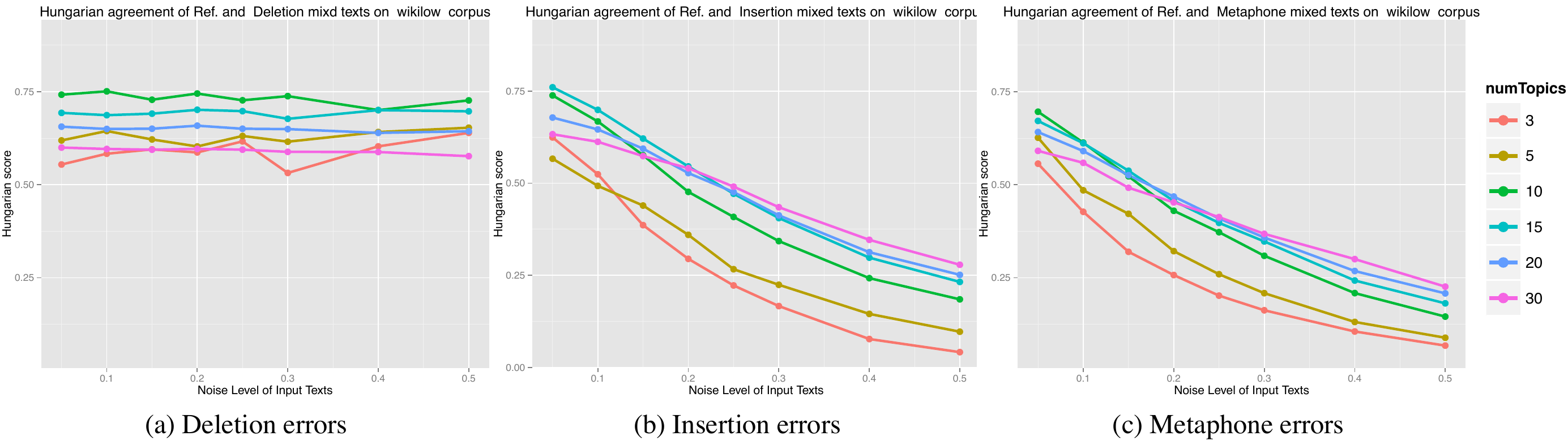}
  \caption{LDA Hungarian scores against noise levels in \textit{wikilow} corpus (10 topics in reference)}
\end{figure*}

Figure 1 and Figure 2 show the topic stability of the $bbc$ and $wikilow$ corpora with reference numbers of topics 5 and 10 each. For each level of topic model complexity, a downward slope indicates decreasing stability of topic models against increasing noise.

In \textit{bbc} corpus, topic model stability shows clear difference with different noise types. The model is especially robust against \textit{Deletion} errors. When noise increases from $5\%$ to $50\%$, the Hungarian agreement score of output topics only drops about 1\% (for the fitted model with $K=5$ in Figure 1(a)). Checking each model in Figure 1(a), We can say that in \textit{bbc} corpus the topic models are robust against random \textit{Deletion} noise. 

In Figure 1(b), the model with 5 topics achieves the highest Hungarian agreement score at noise level 5\% and 10\%, but it drops significantly afterwards. The best and most stable topic model with noise higher than 15\% of \textit{Insertion} errors is the model with 15 topics, which is three times of the reference. Similar trend is observed with \textit{Metaphone} replacement errors in Figure 1(c). The topic model with reference number of topics achieves the highest stability when noise level is low. 
 However, there are differences between Insertion and Metaphone errors in \textit{bbc} corpus tests. With 50\% of \textit{Insertion} errors, the  model with 15 topics achieves 56.4\% agreement with original model, but the agreement is only 32.4\% with \textit{Metaphone} errors. In \textit{bbc} corpus \textit{Metaphone} errors are the most challenging case.

In \textit{wikilow} corpus we observe similar trends in Figure 1 and Figure 2 on specific types of noise. With \textit{Deletion} errors, the topic model with reference number of topics is most stable across noise levels. The difference of topic agreement scores is below 2\% across noise levels. With \textit{Insertion} and \textit{Metaphone} errors, the topic model with reference number of topics is almost the best when noise is low but it drops below others when noise is higher than 15\%. 

Although there are many similarities between Figure 1 and 2, we like to mention two major differences across corpora. In Figure 2(b) and 2(c), Hungarian scores of different topic models (number of topics) share similar gradient of descending slope. However a few models from \textit{bbc} corpus ($K$ as 15, 20, 30) keep roughly stable Hungarian scores in Figure 1(b).
Another difference is that the most stable topic models against noise levels higher than 20\% in Figure 1(b) and 1(c) both have 15 topics, whereas the most stable models in \textit{wikilow} have 30 topics in the same settings. However, if we compare them with corresponding reference topic numbers $K$, the most stable topic models with high systematic errors all have $K*3$ topics. Models with topic number higher than $K*3$ are not optimal in Figure 1(b) and 1(c).

\subsection{Discussion}
\label{sec:discussion}

In Section \ref{sec:lda_output} we observe topic model stability in two corpora and three types of noise. Here we can define a single measurement of topic stability across different settings. If a level of agreement is set as 70\%, LDA is robust against Deletion noises up to 50\% in both \textit{bbc} and \textit{wikilow} corpora. However,  LDA model reaches this agreement level only on 10\% Insertion noises and on 5\% Metaphone replacement noise. We see that Metaphone replacement and insertion are more severe challenges to topic models vs. deletion. 

Regarding deletion errors, we observe that the robustness of a topic model is mostly determined by the number of topics. When this matches the reference, the topic model is the most stable. However, this does not emerge with insertion and metaphone errors. Reference topic models with 5 (bbc) and 10 (wikilow) topics achieve high stability only when noise is $\leq$ 10\%. With higher levels of noise, a more complex topic model exhibits higher stability. 15 (bbc) and 30 (wikilow) topics are the most robust at noise level 50\%. 

A tentative explanation of the high stability of topic models against \textit{Deletion} error concerns the LDA model. LDA takes term frequency into account. The probability of a word belonging to a topic is high if it appears frequently in one topic and seldom in other topics. Such a word is very likely to be an entry in a topic model. If we randomly delete corpus terms, the scale of frequent terms is influenced trivially and these frequent terms still have a high probability of selection.  All rare terms may be removed by deletion, but they have a low chance of appearing in the original topic model anyway. Therefore LDA model has high stability over various levels of deletion errors.
\textit{Insertion} and \textit{Metaphone} replacement introduces systematic noise, which changes the distribution of original texts with respect to frequency, thus having more impact on the LDA model. A high portion of general frequent terms may dilute the frequency of characteristic terms and add noisy terms to a topic model. However, a topic model with many more topics than the reference can deal with the effect of systematic errors.

\section{Conclusions}
\label{sec:conc}

We investigated how transcription errors affect the quality and robustness of topic models produced over a range of corpora, using a topic stability measure introduced \textit{a priori}. We simulate transcription errors from the perspective of word error rate and generated noisy corpora with \textit{deletion}, \textit{insertion} and \emph{Metaphone} replacement. Topic models produced by LDA show high tolerance to deletion noise up to 50\% but low tolerance to insertion and metaphone replacement errors.

We find the robustness of topic models is mainly determined by the extent to which the distribution of original texts is modified. Deletion noise is introduced randomly and its effect on topic models is minor. Insertion and metaphone replacement noise is systematic and undermines topic model stability to a large extent.

Moreover, the number of topics selected also affects topic agreement. With random noise or low-level systematic errors (below 20\%), a correct or approximately correct number of topics brings the highest topic agreement scores. With high level systematic errors, topic models with 3 times the correct number of topics are most robust. In some corpora, redundant number of topics helps the LDA model through severe systematic errors (Figure 1(b)).  This complements previous work by \newcite{bib:Greene14topics} who investigated how topic stability is influenced by number of topics over noise-free corpora.

This suggests that transcribers should perhaps consider omitting words when the uncertainty is high. The topic model is less influenced with a random missing term than an erroneous replacement. For human consumption this may not be optimal, but in the case of output specifically intended for topic extraction this approach makes sense. 

\bibliographystyle{acl}
\bibliography{sigproc} 

\end{document}